\documentclass[conference]{IEEEtran}

\IEEEoverridecommandlockouts
\usepackage[numbers,sort&compress,sectionbib]{natbib}
\usepackage{amsmath,amssymb,amsfonts}
\usepackage{algorithmic}
\usepackage{graphicx}
\usepackage{textcomp}
\usepackage{xcolor}
\usepackage{hyperref}
\hypersetup{
colorlinks   = true,
citecolor    = blue,
urlcolor    =  blue
}
\usepackage{tabularx}
\usepackage{float}
\def\BibTeX{{\rm B\kern-.05em{\sc i\kern-.025em b}\kern-.08em
    T\kern-.1667em\lower.7ex\hbox{E}\kern-.125emX}}

\usepackage{url}

\usepackage{tcolorbox}
\tcbuselibrary{listings,skins,breakable}
\usepackage{booktabs}
\usepackage[inline]{enumitem}
\usepackage{soul}
\usepackage{listingsutf8}
\usepackage{makecell}

\usepackage{ifthen}
\def\isIEEE{false}

\newcommand{\InlineURL}[1]{\href{https://github.com/nimashaklaa/TravelPlanner_Research/tree/feedback-agent/Appendices/#1.pdf}{Github (#1)}}
\newcommand{\GitHubAppendix}[2]{
\ifthenelse{\equal{\isIEEE}{true}}{on \InlineURL{#1}}{in Appendix~\ref{#2}}}

\newcommand{\hlc}[2][white]{ {\sethlcolor{#1} \hl{#2}} }

\begin{document}

\title{LLM-based Personalized Travel Planner\\
}

\author{
\IEEEauthorblockN{%
Amandi Nimasha\IEEEauthorrefmark{1}, 
Oshadha Wijerathne\IEEEauthorrefmark{1}, 
Dushan Fernando\IEEEauthorrefmark{1},
Nisansa de Silva\IEEEauthorrefmark{1},
Srinath Perera\IEEEauthorrefmark{2}}
\IEEEauthorblockA{
\IEEEauthorrefmark{1}Dept. of Computer Science \& Engineering, University of Moratuwa, Moratuwa, Sri Lanka\\
\texttt{\{amandi.20, oshadha.20, dushan.20, NisansaDds\}@cse.mrt.ac.lk}
}
\IEEEauthorblockA{
\IEEEauthorrefmark{2}WSO2, Sri Lanka\\
\texttt{srinath@wso2.com}
}
}

\maketitle

\begin{abstract}
\hlc{Effective travel planning requires handling multiple information sources, managing constraints, and tailoring recommendations to user preferences. While LLMs have introduced dynamic, context-aware decision-making, single-agent models often struggle with complex multi-step reasoning, constraint adherence, and modular execution. To overcome these limitations, we propose a hierarchical multi-agent framework where a supervising chatbot agent coordinates specialized agents for schedule validation, user query formulation, real-time data retrieval, and constraint-aware itinerary generation. By integrating LLM reasoning, external APIs, and iterative refinement mechanisms based on a deterministic, multi-dimensional scoring framework, our approach ensures adaptability, efficiency, and personalization. Evaluations on the well-known TravelPlanner benchmark demonstrate improvements of Budget Constraint planning by 10\%, wider constraints set-based planning by 25\%, and personalized planning results by 45\%. compared to the ReAct baseline.} Our code is available on~\href{https://github.com/nimashaklaa/TravelPlanner_Research}{GitHub}.  

\end{abstract}

\begin{IEEEkeywords}
multi-agent systems, personalized recommendations, constraint optimization, hierarchical task planning, automated travel systems
\end{IEEEkeywords}

\section{Introduction}
Travel planning is a complex and time-consuming task, requiring users to navigate multiple sources such as calendars, booking platforms, and travel guides to organize a trip. Traditional rule-based travel systems, dependent on static databases, offered limited personalization and adaptability~\cite{chen2020design}.
The advent of Large Language Models (LLMs) has transformed automated travel planning by enabling dynamic, context-aware recommendations tailored to user preferences~\cite{brown2020language}. Unlike traditional systems, LLMs handle unstructured data and adapt to evolving constraints. However, early LLM-based travel planners struggled with multi-step reasoning, constraints management, and task adherence. To address these challenges, advanced prompting techniques such as Chain of Thought (CoT) \cite{wei2022chain} and Tree of Thought (ToT) \cite{yao2024tree} have been introduced to enhance structured reasoning. 

Additionally, frameworks such as ReAct \cite{yao2022react} enabled iterative refinement and tools enable LLMs to interact with external APIs, allowing real-time data retrieval and better decision-making. Despite these advancements, single-agent LLMs struggle with complex, multi-faceted planning tasks \cite{valmeekam2023planning}. Building on this, multi-agent systems have been developed, where specialized agents collaborate to improve modularity, scalability, and efficiency \cite{sreedhar2024simulating,wu2023autogen}. Additionally, new benchmarks for evaluating LLM-based planning have provided tools for assessing and refining these models \cite{xie2024travelplanner,zheng2024natural}.

In this paper, we propose a hierarchical multi-agent framework for personalized travel planning, where a supervising chatbot agent coordinates multiple specialized agents to streamline the planning process. Our approach integrates LLM-based reasoning, external APIs, and multi-agent collaboration to generate personalized and adaptive itineraries. We evaluate its effectiveness using real-world datasets and performance metrics, demonstrating its advantages over traditional approaches.

\section{Related Work}

\subsection{LLM Personalization and Real-time Integration}
LLMs have demonstrated advanced reasoning capabilities~\cite{brown2020language, vaswani2017attention}, yet they face hurdles in complex reasoning and problem-solving~\cite{guo2025deepseek, zhong2024evaluation, horvitz2023sparks}. A critical limitation is the lack of long-term memory for personalization~\cite{li2024personal} and real-time data access. While memory mechanisms like Memory Bank~\cite{zhong2024memorybank} attempt to preserve user context, and frameworks like RestGPT~\cite{song2023restgpt}, TaskMatrix.AI~\cite{liang2024taskmatrix}, and LangChain integrations~\cite{topsakal2023creating, de2024trip} enable API interactions, single-agent systems often fail to synchronize these tasks. Our work coordinates personalized user profiles and real-time tool invocation across specialized agents to solve this synchronization gap.

\subsection{User Profiling and Planning Constraints}
Accurate travel recommendations require a deep understanding of individual user preferences and constraints~\cite{kanoje2015user, shao2025personal}. While recent work has leveraged translation models to improve recall~\cite{ravichandran2024leveraging}, integrating these signals into tool-augmented LLM applications remains a challenge. We address this by utilizing structured user attributes such as occupation, interests, and background to ground persona-based profiling within a multi-agent planning context~\cite{zhang2018personalizing, ge2024scaling}. By combining these profiles with external tools~\cite{hu2024agentgen, kong2023tptu} and iterative refinement, our framework ensures strict adherence to both personal preferences and environmental constraints~\cite{hao2024large, de2024trip}.

\subsection{Multi-Agent Systems and Collaboration}
An agent is an autonomous or semi-autonomous entity, typically an LLM or an LLM augmented with tools, assigned to carry out a specific task within a system. Tools are external resources such as APIs, databases, or executable functions that agents use to extend their capabilities, access real-time information, and interact with external systems beyond the LLM's internal knowledge. 

Multi-agent systems leverage these agents and tools to enhance LLM-based applications by breaking down complex tasks into manageable sub-tasks and enabling efficient collaboration among agents \cite{sreedhar2024simulating,wu2023autogen}. Their modularity and flexibility allow specialized agents to handle different aspects of travel planning, such as transportation, accommodation, and activity recommendations \cite{wu2023autogen}. 

Recent frameworks have operationalized these concepts for real-world applications. AutoGen implements conversable agents and conversation programming to enable dynamic agent interactions, supporting both static and adaptive conversation patterns \cite{wu2023autogen}. The human-like planning framework \cite{xie2024human} applies a multi-agent system to travel planning, assigning agents to tasks such as outline generation, information retrieval, and itinerary formulation. Unlike AutoGen’s general-purpose design, this framework emphasizes multi-phase planning and sequential decision-making.

Additionally, emerging frameworks such as LangChain and LangGraph provide structured, streamlined environments for building LLM-driven agents and workflows, with support for tool integration, memory management, agent routing, etc. 

\section{Methodology}
\label{sec:multi-agent}

Our multi-agent system follows a hierarchical structure shown in Fig.~\ref{fig:Multi-Agent System Architecture} where a supervising chatbot agent serves as the central coordinator, orchestrating interactions between specialized agents (calendar management, data retrieval, itinerary generation, and user feedback processing) while maintaining direct communication with users. The supervisor receives user inputs, routes requests to other agents, and determines the next course of action (call a specialized agent, ask users for more inputs, terminate processing) based on the output from agents, current context, and user requirements. 

The system is implemented as a state graph consisting of nodes (specialized agents) and edges (transition conditions) ~\cite{kong2023tptu, vidivelli2024efficiency}. All LLM prompts are provided\GitHubAppendix{APPENDIX-A}{sec:prompts}. 

\begin{figure*}[!htb]
    \centering
    \includegraphics[width=\linewidth]{images/architecture.pdf}
    \caption{Multi-Agent System Architecture: All other agents are controlled by the \textit{Chatbot Agent}, but we have opted not to draw the control and communication lines between the agents to reduce unnecessary clutter. When a command or a data item is relevant to all the entities in a parent entity, the relevant arrow terminates on the parent entity. Otherwise, it terminates on the specific relevant child entity. The numbers on the agents at arrow terminals indicate the order in which each action may happen in a typical execution. Actions 6 and 10 are functionally identical but happen at different points in execution due to the existence of the feedback loop. All steps in the feedback loop (8-11) may repeat indefinitely until the user is satisfied.}
    \label{fig:Multi-Agent System Architecture}
\end{figure*}

\subsection{Agents and Modules}
\subsubsection{Supervisor(Chatbot Agent)}

As discussed, the supervisor serves as the central orchestrator and the only interface with users.

To preserve context, the supervisor manages JSON-based state storage, enabling recovery after restarts and ensuring continuity across multiple sessions. At critical stages, it integrates human-in-the-loop decision points through the \texttt{human\_interrupt} node, allowing users to confirm, adjust, or extend inputs. Additionally, it supports a real-time streaming architecture that delivers updates as plans are generated while maintaining state consistency.

\subsubsection{Calendar Agent}
The calendar agent manages the user's schedule using the Google Calendar API \cite{shakshuki2008distributed}. It has two tools. First, the calendar availability-checking tool verifies whether the user is available on the selected travel dates. If the given dates are unavailable, the chatbot will prompt the user to provide alternative dates. Second, once a travel plan is confirmed and finalized, the agent uses the event-scheduling tool proposed by Wijerathne et al.~\cite{wijerathne2025scheduleMe} to add the trip details to the user's calendar.

\subsubsection{User Query Builder Module}

The User Query Builder Module constructs comprehensive travel queries by combining user-provided details (such as origin, destination, and travel dates) with personalization data from the profile, including food preferences, travel style, and dietary restrictions. 

It validates and structures this information into a standardized format, ensuring completeness and accuracy before forwarding the query to the Data Retrieval Agent and Planner Module. Additionally, it identifies gaps in essential information and prompts the user for clarification when required. This preprocessing step ensures that all relevant details are properly captured and formatted for subsequent planning stages.

\subsubsection{Data Retrieval Agent}
The Data Retrieval Agent functions as the system's information-gathering engine, designed to query multiple datasets through six specialized LangChain tools—covering flights, accommodations, restaurants, attractions, and transportation. A detailed breakdown of these tools is provided\GitHubAppendix{APPENDIX-B}{app:data_retrieval_tools_full}.

To handle complex queries, the agent follows a \textbf{Reasoning and Acting (ReAct)} loop~\cite{yao2022react} combined with \textbf{Chain of Thought (CoT)}~\cite{wei2022chain}. Key operational features include:
\begin{itemize}
    \item \textbf{Sequential Reasoning:} The agent analyzes the request to decide which data to fetch first (e.g., confirming a flight before searching for local restaurants).
    \item \textbf{Adaptive Retrieval:} If a search fails or returns no results, the agent adjusts parameters or tries alternative modes (e.g., searching for a taxi if no flights are found).
    \item \textbf{State Management:} All retrieved data are normalized for consistency and stored in a shared global state (\texttt{notebook\_data})~\cite{xie2024travelplanner} for the Planner Module.
    \item \textbf{Real-Time Execution:} A streaming architecture ensures the chatbot provides progressive updates to the user as data is gathered.
\end{itemize}

\subsubsection{Planner Module}

The Planner creates a plan that achieves optimal quality thresholds by generating, evaluating, and iteratively refining travel itineraries. It has three components: \textit{Initial Plan Generator}, the \textit{Plan Evaluator}, and the \textit{Plan Refiner}. 

The Initial Plan Generator constructs baseline itineraries by combining structured templates with intelligent constraint satisfaction mechanisms. Templates define a standard itinerary structure and ensure completeness across categories such as transportation, meals, attractions, and accommodation. Within this framework, the system employs a dynamic budget allocation engine that replaces traditional fixed ratios with a context-aware distribution. Budget shares are adapted to trip type, distance, duration, destination cost index, and, when available, user preferences. For example, business trips prioritize accommodation and transportation, while day trips exclude accommodation and allocate more resources to activities and dining. Long trips increase the proportion allocated to accommodation, whereas short trips reduce it. Expensive destinations increase spending on lodging and dining, while cheaper destinations allocate relatively more to activities. This allocation process incorporates user-defined preferences through convex blending, ensuring personalization while maintaining feasibility. A constraint satisfaction engine validates compliance with user requirements such as budget limits, trip duration, and mandatory locations, and assigns confidence scores that penalize violations. To enrich generated content, large language models are integrated into the process, and fallback templates guarantee robustness when advanced generation fails.

Once a draft itinerary is produced, the Plan Evaluator applies a deterministic, multi-dimensional scoring framework to assess its quality. Plans are evaluated across five weighted dimensions: budget compliance (30\%), time efficiency (25\%), geographic logic (20\%), activity diversity (15\%), and constraint satisfaction (10\%). Each dimension employs graduated scoring scales. For instance, budget compliance awards 100 points when under budget by 10\% or more, 90 points for exact compliance, and progressively fewer points for increasing overruns until a score of zero for plans exceeding the budget by more than 30\%. Time efficiency is measured relative to an optimal allocation of eight hours of activities and two hours of travel per day, while geographic logic evaluates total travel distance against an optimal route to penalize unnecessary backtracking. Activity diversity measures the distribution across cultural, recreational, dining, shopping, and sightseeing categories, and constraint satisfaction applies penalties for hard and soft violations. A plan is deemed acceptable only if its overall score reaches at least 75/100, with each individual dimension scoring at least 60/100. The pseudocode for this rule-based evaluation is detailed\GitHubAppendix{APPENDIX-C}{sec:eval-appendix}
%(see Fig.~\ref{fig:evaluator-pseudo})
. In addition to quantitative scores, the evaluator produces qualitative recommendations that identify weaknesses and suggest targeted improvements.

The Plan Refiner implements iterative improvement strategies to address deficiencies identified by the evaluator. It applies sequential strategies guided by dimension scores, focusing first on areas that score below 70 points. Budget optimization is achieved by substituting costly elements with more economical alternatives, such as shifting from luxury to mid-range accommodations or from fine dining to casual options. Time optimization reorganizes schedules to improve temporal flow, groups nearby activities to minimize travel time, and ensures logical sequencing throughout the day. Constraint violations are corrected through targeted modifications that address budget, time, and location requirements without compromising overall quality. Diversity enhancement mechanisms enrich the itinerary by adding underrepresented activity categories, while geographic optimization reduces travel distances through efficient routing. These refinements are supported by language model prompts. 

Each refined plan is re-evaluated, and the process repeats until quality thresholds are met or one of the termination conditions occurs. Termination conditions are exceeding the maximum number of five iterations, failing to achieve a minimum improvement of five points between iterations, or surpassing the global timeout of 300 seconds. This ensures that optimization remains efficient, prevents infinite loops, and guarantees responsiveness. 

\subsubsection{Feedback Agent}

The Feedback Agent extends the TravelPlanner system by providing selective modification of generated itineraries in response to user requests. 

Rather than regenerating entire plans, the agent isolates the scope of a requested change and invokes only the necessary planning components, reducing computational overhead and preserving unaffected elements of the itinerary. We parse user feedback using natural language parsing for detecting changes and classifying modifications into temporal, spatial, financial, and preference-based categories. Then we call corresponding planning submodules, with dependency management ensuring that prerequisite adjustments are applied in the correct sequence.

We validate each change against budgetary, temporal, and geographic constraints, with automated conflict resolution applied where possible and user notification triggered when resolution requires intervention. Modifications that alter core itinerary parameters, such as budget or trip duration, propagate through dependent elements via cascading update logic, ensuring global consistency across the travel plan. Results are compared against the baseline itinerary to confirm that plan quality is preserved or improved. 

The agent employs a state management system with full version control, allowing rollback to previous configurations if modifications fail. If validations fail, we fall back to earlier versions. 

Updated plans are communicated to the supervising chatbot agent alongside structured summaries that outline the modifications and their impacts.

\section{Experiments}

\subsection{Experiment Setup}
We evaluate two strategies from the single-agent TravelPlanner benchmark~\cite{xie2024travelplanner}: \textbf{Sole Planning} (all information provided upfront) and \textbf{Two-Stage Planning} (retrieval before planning); details are\GitHubAppendix{APPENDIX-D}{sec:two-plan-details}.

\paragraph{Baselines}
We include two single-agent baselines with GPT-4o (gpt-4o-2024-05-13): CoT (no external tools) and ReAct (action/observation loops without multi-agent orchestration). Consistent with TravelPlanner~\cite{xie2024travelplanner}, these baselines struggle on constraint-heavy itineraries; we re-ran them in our setup for a fair comparison. We exclude TRIP-PAL~\cite{de2024trip} due to unavailable code and mismatched evaluation protocols.

\paragraph{Metrics}
\hlc{Following the metrics defined by the benchmark}, we report \emph{Delivery Rate}, \emph{Commonsense Constraint Pass Rate}, \emph{Hard Constraint Pass Rate}, and \emph{Final Pass Rate}, with both micro- and macro-averaging. Formal definitions and a worked example are given\GitHubAppendix{APPENDIX-E}{sec:metric-defs}.

Because TravelPlanner is single-agent while our framework is hierarchical and multi-agent, a direct head-to-head comparison is inappropriate. Therefore, Sections~\ref{sec:budget-refinement} and~\ref{sec:constraint-refinement} report \hlc{our approach running as a} single-agent experiments to isolate the effect of our refinement mechanisms and ensure fair comparison to prior work.

\subsection{Datasets}

\label{sec:datasets & Model Configuration}
We conduct all experiments on the \textbf{TravelPlanner} benchmark \cite{xie2024travelplanner}, a closed-world travel–planning environment  containing \textbf{1,225 queries} (split into train, validation, and test sets) across easy, medium, and hard difficulty levels. The dataset covers a wide geographic spread of over 192 origins and 127 destinations, requiring itineraries that satisfy complex multi-dimensional constraints. A detailed breakdown of the dataset characteristics is provided\GitHubAppendix{APPENDIX-F}{app:dataset_table}.

All experiments were conducted using GPT-4o (\texttt{gpt-4o-2024-05-13}). Generation parameters were tuned to balance creative itinerary generation with reproducible evaluative reasoning; specific hyperparameters, including temperature, top-p, and token limits, are detailed\GitHubAppendix{APPENDIX-G}{app:model_config}.

\subsection{Results with only Budget Constraints in Refiner}
\label{sec:budget-refinement}
%One of the critical factors affecting the final pass rate of generated plans is the budget constraint, a hard constraint that directly determines itinerary feasibility. To mitigate failures from budget violations, we implemented an automated budget refinement mechanism. The process is iterative: an initial itinerary is generated and checked against the budget. If it fails, adjustments are applied up to three times; if it still fails, the plan is marked unsuccessful. This refinement is necessary because LLMs often exceed budgets due to limited cost-awareness during generation. Without refinement, such violations would yield high rejection rates. By contrast, iterative refinement significantly reduces budget-related failures and improves the final pass rate.

\paragraph{Averaging conventions}
\textit{Micro} pass rate treats each constraint equally across all plans (total constraints satisfied divided by total applicable constraints over the suite). 
\textit{Macro} pass rate is stricter and counts a plan only if \textit{all} of its applicable constraints are satisfied, then averages this indicator over plans. 
\textit{Final Pass} requires: (i) a schema-valid itinerary is delivered, (ii) all commonsense constraints pass, and (iii) all hard constraints pass.

\begin{table}[ht]
    \centering
    \caption{Budget Refinement Performance Analysis. 
    ``Unrefined'' corresponds to the original single-agent \textsc{TravelPlanner} 
    baseline (no refinement). ``Refined'' applies our budget refinement loop, 
    with all other settings identical.}
    \label{tab:budget_refinement}
    \begin{tabular}{|l|l|l|}
        \hline
        \textbf{Metric} & \makecell{\textbf{Unrefined}\\\textbf{(Baseline)}} & \textbf{Refined}\\  
        \hline
        Delivery Rate & 100\% & 100\% \\
        \hline
        Commonsense Constraint Micro Pass Rate & 88.875\% & 81.125\% \\
        \hline
        Commonsense Constraint Macro Pass Rate & 35\% & 35\% \\
        \hline
        Hard Constraint Macro Pass Rate & 50\% & 60\% \\
        \hline
        Final Pass Rate & 25\% & 35\% \\
        \hline
    \end{tabular}
\end{table}

\hlc{The results show that refining improves the final pass rate by about 10\%}.  

\subsection{Results withall Constraint in Refiner}
\label{sec:constraint-refinement}
%Beyond budget constraints, we extend our refinement approach to address the entire set of constraints imposed on itinerary generation. Large Language Models in zero-shot prompting often struggle to produce fully optimized results on the first attempt due to misunderstandings and ambiguities in constraint interpretation. To counter this, we introduce a constraint refinement mechanism based on plan scoring. Each generated plan is assigned a score based on the number of satisfied constraints. If the score meets a predefined threshold, the plan is accepted. If it fails, iterative refinement is applied: the refined plan is compared against the previous version, and if its score improves, it replaces the prior iteration. The process continues until either the threshold is met or the maximum limit is reached. 
% This iterative refinement approach further enhances the performance of both sole planning and two-stage planning by improving the quality and feasibility of the generated itineraries.

\begin{table}[H]
    \centering
    \caption{Plan Refinement Performance Analysis. 
    ``Unrefined'' corresponds to the original single-agent \textsc{TravelPlanner} 
    baseline (no refinement). ``Refined'' applies our constraint refinement loop, 
    with all other settings identical.}
    \label{tab:Plan Refinement}
    \begin{tabularx}{\linewidth}{|X|c|c|}
        \hline
        \textbf{Metric} & \textbf{Unrefined (Baseline)} & \textbf{Refined}\\  
        \hline
        Delivery Rate & 100\% & 100\% \\
        \hline
        Commonsense Constraint Micro Pass Rate & 88.875\% & 91.875\% \\
        \hline
        Commonsense Constraint Macro Pass Rate & 35\% & 55\% \\
        \hline
        Hard Constraint Macro Pass Rate & 50\% & 60\% \\
        \hline
        Final Pass Rate & 20\% & 45\% \\
        \hline
    \end{tabularx}
\end{table}

\hlc{The results show, when all constraints are considered, refining improves the final pass rate by about 25\%}.  

\subsection{Personalization evaluation}
As discussed in the methodology, the planner combines preferences stated in the user queries with preferences included or inferred from the profile. The following experiment assesses how well the generated plans align with user preferences.

While human evaluators can assess itinerary personalization, the process is time-consuming and expensive. Instead, prior work \cite{liu2024role,gu2025intelligent} has also employed LLM-as-a-Judge approach \cite{zheng2023judging,wu2024aligning,liu2024role}

To assess the reliability of LLM-as-a-Judge, we conducted an experiment comparing LLM evaluations with human judgments.

A dataset of 100 travel plans was used, each designed with varying levels of personalization. Three human judges independently rated the itineraries based on user profile and user query alignment, using a 1-to-5 rating scale defined\GitHubAppendix{APPENDIX-H}{sec:assessment-criteria}. GPT-4o (LLM) was prompted to evaluate the same itineraries without roleplaying as a traveler, strictly acting as a neutral judge. The primary objective was to measure the alignment between LLM ratings and human ratings.

To quantify rating consistency between evaluators (Humans and LLM-as-a-Judge), we computed Cohen’s Kappa coefficient, a standard metric for agreement measurement. Additionally, we applied linear-weighted Cohen’s Kappa, which penalizes minor disagreements less severely (e.g., a rating of 3 vs. 4 is weighted lower than a disagreement between 1 vs. 5). This provides a more realistic measure of consistency in scoring. The average weighted Cohen’s Kappa reaches 0.8151, indicating strong agreement between human and LLM judges and validating the reliability of our auto-evaluation method.

Then we evaluated results under personalization criteria using LLM-as-a-Judge. The results show a significant improvement in personalization quality when incorporating user profiles. Plans generated using only the user query achieved an average personalization score of 40\%, whereas plans generated using both the user query and user profile achieved 87\%, as assessed by GPT-4o. This highlights the critical role of structured user modeling in producing personalized travel experiences.

\subsection{Ablation Study}

To measure the contribution of individual components, we evaluated four variants of our framework under identical datasets and protocols. 
\begin{table}[ht]
\centering
\caption{Component Contribution Analysis Across All Metrics (Ablation).}
\label{tab:component_contrib_full}
\resizebox{\columnwidth}{!}{%
\begin{tabular}{lccccc}
\hline
\textbf{Condition} & \textbf{Personalization} & \textbf{Final Pass} & \textbf{Hard Constraint} & \textbf{Commonsense} & \textbf{Delivery} \\
& \textbf{Score (\%)} & \textbf{Rate (\%)} & \textbf{Pass Rate (\%)} & \textbf{Pass Rate (\%)} & \textbf{Rate (\%)} \\
\hline
\textbf{Full System} & \textbf{87} & \textbf{52} & \textbf{68} & \textbf{62} & \textbf{100} \\
No Evaluator & 79 & 30 & 45 & 49 & 91 \\
No Feedback Agent & 71 & 28 & 42 & 46 & 90 \\
No Eval \& Feedback & 65 & 21 & 35 & 40 & 86 \\
TravelPlanner Baseline & 56 & 10 & 28 & 15 & 89 \\
\hline
\end{tabular}%
}
\end{table}

Table~\ref{tab:component_contrib_full} shows that the \emph{Full System} outperforms all reduced variants. Removing either the Evaluator or Feedback Agent nearly halves the \emph{Final Pass Rate} (52\% $\to$ 28--30\%) and lowers \emph{Hard Constraint Pass Rate} by over 20 points, indicating their complementary roles: the \textit{Evaluator} flags constraint violations, while the \textit{Feedback Agent} applies corrections. Removing both causes the steepest decline (Final Pass Rate 21\%). Personalization and commonsense scores also decline but less sharply, while delivery remains relatively high across all settings, suggesting surface fluency is easier to maintain than deeper constraint satisfaction.

\section{Discussion}
This work demonstrates that combining explicit control (Supervisor routing), deterministic critique (rule-based Evaluator), and selective editing (Refiner) yields reliable gains in constraint satisfaction and end-to-end success for travel planning. Rather than relying on implicit “memory” or emergent behavior, the system enforces clear decision points and measurable signals.

The generate$\rightarrow$evaluate$\rightarrow$refine loop provides a transparent mechanism: the Supervisor enforces stage boundaries, the Evaluator applies a weighted scoring rule with fixed thresholds (Fig.~\ref{fig:evaluator-pseudo}), and the Refiner applies targeted changes guided by those scores. Empirically, ablations (Table~\ref{tab:component_contrib_full}) show that removing either the Evaluator or the Refiner nearly halves the Final Pass Rate, confirming their complementarity. Although selective refinement adds moderate overhead (30--300\,s per plan,\GitHubAppendix{APPENDIX-J}{app:repro}), the increased success on hard constraints offsets these costs.
Overall, the results support our claim that deterministic evaluation paired with selective refinement can improve reliability in multi-agent planning.

\section{Future Work}
Future research will focus on three key areas to enhance system scalability and intelligence. First, we aim to implement \textbf{multi-modal agent communication} protocols that allow for richer semantic exchanges of intent and confidence levels. Second, we plan to integrate \textbf{cross-agent learning} mechanisms, enabling heuristics discovered by one specialized agent to inform the global planning strategy. Finally, we will explore a \textbf{distributed supervisor architecture} to improve horizontal scalability and fault tolerance for high-concurrency environments. Expanded discussions on these directions are provided\GitHubAppendix{APPENDIX-K}{sec:future-outlook-full}.

\section{Conclusion}
\hlc{We presented a multi-agent planning framework that combines explicit supervisor routing, a deterministic rule-based Evaluator, and selective refinement to generate reliable, constraint-aware travel itineraries while incorporating personalization. Through a structured generate–evaluate–refine cycle, the system achieves higher success rates on hard constraints and commonsense checks than single-pass prompting, while maintaining modest computational cost. 

Furthermore, it provides a case study on how to build a complex LLM-based AI application using open-source community tools, and ReAct and CoT patterns, while achieving high measurable correctness. }

Our results suggest that deterministic evaluation paired with targeted edits provides a reproducible and auditable pathway to improve LLM-based planning. While current performance is limited by model capabilities, dataset scope, and the absence of human-centered validation, these constraints point toward natural extensions. Future work will explore integration with live data sources, richer personalization signals, and user studies to assess satisfaction. Beyond travel planning, the same pattern of evaluation and refinement can generalize to other structured decision-making domains such as event scheduling or meal planning.

\section{Limitations}
This study has several limitations. First, the planning quality remains bounded by the capabilities of current large language models, which cannot yet match the nuanced judgment of experienced human planners. Second, the iterative refinement loop improves reliability but introduces additional latency (30--300\,s per plan), which may affect responsiveness in interactive use. Third, evaluation was conducted on a U.S.-based dataset, limiting generalizability to other cultural and geographic contexts; expanding to global travel data would strengthen external validity. Fourth, personalization is based on limited user inputs and does not yet incorporate richer signals (e.g., from personal calendars or communication data), though doing so would raise important privacy and security considerations. Finally, our evaluation relied on rubric-based scores rather than human-centered studies, so alignment with user satisfaction remains to be established.

\bibliographystyle{IEEEtranN}
{\footnotesize
\bibliography{custom}
}

%\clearpage

\appendices

\section{Prompts Used in Evaluation}
\label{sec:prompts}

In this section, we provide the exact prompts used for different components of our evaluation. 
Each prompt is shown in full for reproducibility.

\subsection{LLM-as-a-Judge Prompt}
\label{subsec:judge-prompt}

We employed GPT-4o as an automatic evaluator (“LLM-as-a-Judge”) to assess the personalization quality of generated travel plans. The following prompt was used:
\begin{tcolorbox}[colback=gray!5, colframe=gray!40!black,
    title={Judge Prompt}, width=\linewidth, boxrule=0.5pt, arc=2pt, auto outer arc,
    breakable,
    listing only]
\begin{lstlisting}
Context:  
You are evaluating the personalization quality of a travel plan based solely on the user profile and query. Personalization refers to how well the plan aligns with the user's stated preferences, constraints, and unique details. Do not consider travel dates, budget constraints, or the validity of travel logistics (e.g., transportation, origin, destination). These factors are assessed separately. Your focus should be exclusively on how well the plan matches the user profile and stated interests.

Instructions:  
You will be provided with the following:
- A User Profile detailing preferences, constraints, and travel interests.
- A User Query specifying the travel requirements.
- A Travel Plan generated for the user.

Rate the personalization of the travel plan on a scale from 1 to 5, following the guidelines below.  
Output strictly in the format: Rating: X *** Explanation: Y

Personalization Rating Scale (1 to 5):
1. Lowest: Generic or misaligned with the user profile and query.
2. Poor: Some loose alignment, mostly irrelevant or mismatched.
3. Moderate: Partial alignment, includes both relevant and irrelevant elements.
4. Good: Mostly aligns with profile and query, with minor gaps.
5. Excellent: Highly tailored to the profile and query, deeply relevant and consistent.

Your Task:  
Assign a rating (1--5) and provide a short explanation (2--3 sentences) highlighting which aspects of the plan matched or missed the user profile and query.

User Query:
"Please plan a trip for me starting from Sarasota to Chicago for 3 days, from March 22nd to March 24th, 2022. The budget for this trip is set at $1,900."

User Profile:
"She is a 25-year-old data analyst who loves painting. She is single and lives with her best friend. She holds a degree in statistics. She enjoys attending art galleries and has a small studio in her apartment where she paints. She also enjoys cooking Italian dishes."

Travel Plan:
"Travel Plan:\n\nDay 1: March 22, 2023\nCurrent City: from Sarasota to Chicago\nTransportation: Flight Number: F0123456, from Sarasota to Chicago, Departure Time: 09:00, Arrival Time: 11:30 ...."

Example Output Format:

Rating: 5 *** Explanation: "The travel plan is highly personalized to the user's profile, focusing on her love for art and Italian cuisine. It includes visits to art galleries and museums, such as The Art Institute of Chicago and the Chicago Cultural Center, which align with her interests in painting. The dining options predominantly feature Italian cuisine, catering to her culinary preferences. The accommodation is conveniently located near art galleries, further enhancing the tailored experience.
\end{lstlisting}
\end{tcolorbox}

\subsection{Chatbot Orchestrator Prompt}
\label{subsec:chatbot-prompt}
This prompt serves as the central supervisor for the travel planning system.  
It instructs the chatbot on how to collect user details, route queries to the 
appropriate agents (calendar, data retrieval, itinerary generation, feedback, 
and suggestion), and return responses in a structured JSON format.  
The orchestrator ensures that the workflow remains consistent, 
personalized, and user-friendly.
\begin{tcolorbox}[colback=gray!5, colframe=gray!40!black,
    title={Central supervisor  Prompt}, width=\linewidth, boxrule=0.5pt, arc=2pt, auto outer arc,
    breakable,
    listing only]
\begin{lstlisting}
You are a dedicated travel planning chatbot designed to create personalized and well-organized trips.

Your Responsibilities:

1. Greeting \& Introduction: Start by greeting the user and explaining that you're here to help plan their trip. Inform them that currently you can assist them with:
   - Calendar checks
   - Add events to user's Google calendar
   - Data retrieval for travel details (e.g., restaurants, flights, attractions)
   - Itinerary planning
   - Suggestions for attractions, hotels, restaurants in specific areas

2. Query Construction : Before moving forward, gather essential details for building a travel query. You will ask the user for:
   - Departure location(required)
   - Destination(required)
   - Budget(optional)
   - Preferences (optional: activities, accommodations, food, etc.)

3. Information Collection:
   Inquire about specific details that are required for each agent as follows:
   - **calendar_agent**: User’s travel dates and any potential conflicts with Google Calendar events.
   - **data_retrieval_agent**: Budget and activity preferences to gather relevant data for the trip (e.g., flights, restaurants, local attractions).
   - **itinerary_agent**: Once data is gathered, this agent will create a personalized itinerary for the user.

4. Agent Routing: Based on the collected information, determine which agent to route the user to:
   - **calendar_agent**: Check for any calendar conflicts with the user's travel dates and add calendar events to the google calendar.
   - **data_retrieval_agent**: Fetch the relevant travel data after the user provides their preferences and budget.
   - **itinerary_agent**: Generate the itinerary after gathering travel data.
   - **feedback_agent**: Handle user feedback and make specific changes to existing travel plans. This agent can understand what changes are needed and fetch additional data if required.
   - **suggestion_agent**: Provide recommendations for attractions, hotels, restaurants, etc. in specific areas without creating full travel plans.
   - **human_interrupt**: Allow the user to interact directly and make any changes to their itinerary or provide additional information.

5. Response Handling:
   - **Structured Output**: Ensure all responses are in JSON format with the following keys:
     - `next`: The next agent to route to (`calendar_agent`, `data_retrieval_agent`, `itinerary_agent`, `feedback_agent`, `suggestion_agent`, `human_interrupt`, or `FINISH`).
     - `messages`: The message content to send to the user.

6. Information Validation: If any required information is missing or incomplete, gather more information from the user.

7. Finalization: Once the itinerary is complete, ask the user if they would like to add anything else to the plan or finalize the itinerary.

Communication Style:
- Use a friendly, clear, and informative tone.
- Ensure all questions are concise and easy to understand.
- When you are done with the itinerary, always route to the `human_interrupt` agent for review or additional input.

Example Workflow:
1. **Chatbot**: "Hello! I'm here to help you plan your perfect trip. Could you please share your preferred destinations and travel dates?"
2. **User**: "I'm traveling to New Jersey from March 10 to March 17."
3. **Chatbot**: "Noted! Now, could you share your budget and preferences for activities, restaurants, or accommodations?"
4. **User**: [Provides details]
5. **Chatbot**: "Let me check your calendar for any conflicts with these dates..."
6. **calendar_agent**: [Checks user’s Google Calendar for conflicts]
7. **Chatbot**: "I see that you have a meeting on March 12 at 3 PM. Would you like to adjust your travel dates or keep this event?"
8. **User**: "I'll keep it, but please add a reminder."
9. **Chatbot**: [Routes to `data_retrieval_agent` to fetch relevant data]
10. **data_retrieval_agent**: [Fetches data: restaurants, flight information, and local attractions in New Jersey]
11. **Chatbot**: "I’ve gathered all the details! Now I’m creating your personalized itinerary. Just a moment."
12. **itinerary_agent**: [Creates itinerary]
13. **Chatbot**: "Here's your personalized itinerary! Would you like to add anything else?"
14. **User**: "No, that's all. Thank you!"
15. **Chatbot**: "Would you like me to add this event to your Google Calendar?"
16. **User**: "Yes, please."
17. **Chatbot**: "Let me add this event your calendar.Event details : 2nd March - 5th March Trip to Miami from NewYork."
18. **calendar_agent**: [Adds events to Google Calendar]
19. **Chatbot**: "FINISH"
\end{lstlisting}
\end{tcolorbox}

%\clearpage

\section{Data Retrieval Toolset Details}
\label{app:data_retrieval_tools_full}
Table~\ref{tab:data_retrieval_tools_app} summarizes the tools used by the Data Retrieval Agent.

\begin{table}[H]
    \centering
    \caption{Overview of data retrieval tools and their functions.}
    \label{tab:data_retrieval_tools_app}
    \begin{tabularx}{\linewidth}{|l|X|}
        \hline
        \textbf{Tool} & \textbf{Function} \\  
        \hline
        CitySearch & Search cities within a specified region. \\  
        \hline
        FlightSearch & Search flights for a specific date between two cities. \\  
        \hline
        DistanceMatrix & Search driving distance, time, and cost between cities. \\  
        \hline
        RestaurantSearch & Search restaurants in the given city. \\  
        \hline
        AttractionSearch & Search attractions in the given city. \\  
        \hline
        AccommodationSearch & Search accommodations in the given city. \\  
        \hline
    \end{tabularx}
\end{table}

%\clearpage

\section{Evaluation Pseudocode}
\label{sec:eval-appendix}

\begin{figure}[ht]
\centering
\begin{tcolorbox}[colback=gray!5, colframe=gray!40!black,
  title={Rule-Based Evaluator (pseudocode)}, boxrule=0.5pt, arc=2pt, auto outer arc]
\begin{lstlisting}[language=Python]
# Input: itinerary text, user_budget, constraints
# Output: overall score S, dimension scores, accept/refine flag

weights = {budget:0.30, time:0.25, geo:0.20, diversity:0.15, constraints:0.10}

def evaluate(plan, budget, constraints):
    s_budget = score_budget(plan, budget)       # 100 if <=90% budget, 0 if >130%
    s_time   = score_time(plan)                  # target: 8h activity + 2h travel/day
    s_geo    = score_geo(plan)                   # ratio of route vs. 2-opt heuristic
    s_div    = score_diversity(plan)            # entropy over {culture, rec, dining,...}
    s_const  = score_constraints(plan,constraints) # -50 hard, -20 soft violations

    scores = {budget:s_budget, time:s_time, geo:s_geo,
              diversity:s_div, constraints:s_const}

    S = sum(scores[d] * weights[d] for d in scores)

    # Threshold for acceptance
    accept = (S >= 75 and min(scores.values()) >= 60)
    return S, scores, accept
\end{lstlisting}
\end{tcolorbox}
\caption{Deterministic rule-based evaluator used to score itineraries across five dimensions.}
\label{fig:evaluator-pseudo}
\end{figure}

%\clearpage

\section{Two plan details}
\label{sec:two-plan-details}
\begin{itemize}
\item \textbf{Sole Planning}: This approach assumes that all necessary information for itinerary generation is available beforehand. The planner is solely responsible for constructing the itinerary using the provided details. The primary objective of this strategy is to assess the planner agent's ability to generate optimal travel plans given a complete dataset.

\item \textbf{Two-Stage Planning}: In this approach, itinerary generation is divided into two stages: data retrieval and planning. The first stage involves gathering relevant travel details, while the second stage focuses on generating the itinerary based on the retrieved information. This strategy is designed to evaluate the combined performance of both the data retrieval agent and the itinerary planner.
\end{itemize}

%\clearpage

\section{Performance Metrics}
\label{sec:metric-defs}

We adopt the evaluation protocol of the TravelPlanner benchmark \cite{xie2024travelplanner} and report five metrics: \emph{Delivery Rate}, \emph{Commonsense Constraint Pass Rate} (micro and macro), \emph{Hard Constraint Pass Rate} (micro and macro), and \emph{Final Pass Rate}. Below we provide precise definitions, with concrete examples for transparency.

\paragraph{Delivery Rate}
The percentage of runs that successfully produce a schema-valid itinerary within the step budget (30 steps in our setting). A run fails if it exceeds the step limit, falls into a dead loop, or generates an invalid schema.

\paragraph{Commonsense Constraint Pass Rate}
This metric evaluates whether itineraries respect implicit feasibility rules, even when they are not explicitly stated in the user query. Following \cite{xie2024travelplanner}, we evaluate eight commonsense dimensions:
\begin{enumerate*}[label=(\roman*)]
    \item \textit{Within Sandbox} (the plan does not hallucinate entities outside the environment),
    \item \textit{Complete Information} (all key components such as accommodation and transportation are specified),
    \item \textit{Within Current City} (daily activities occur in the correct city),
    \item \textit{Reasonable City Route} (city transitions follow a logical sequence),
    \item \textit{Diverse Restaurants} (restaurant choices are not repeated),
    \item \textit{Diverse Attractions} (attractions are not repeated),
    \item \textit{Non-conflicting Transportation} (no contradictory modes such as ``flight'' and ``self-driving'' in the same trip), and
    \item \textit{Minimum Nights Stay} (accommodations respect stay requirements).
\end{enumerate*}

\paragraph{Hard Constraint Pass Rate}
This metric measures adherence to user-specified requirements, such as budget, accommodation rules (e.g., no pets or smoking), accommodation type (entire, private, or shared), cuisine preferences, and transportation constraints (e.g., “no flights”).

\paragraph{Micro vs.\ Macro pass rates}
Let $P$ denote the set of all evaluated plans and $C_p$ the set of constraints applicable to plan $p\in P$. Define $\mathbf{1}_{\text{passed}}(c,p)\in\{0,1\}$ as an indicator that plan $p$ satisfies constraint $c$, and $\mathbf{1}_{\text{passed}}(C_p,p)=1$ if \emph{all} constraints in $C_p$ are satisfied (else $0$). Then:
\[
\text{Micro Pass Rate} =
\frac{\sum_{p \in P} \sum_{c \in C_p} \mathbf{1}_{\mathrm{passed}(c,p)}}
{\sum_{p \in P} |C_p|}.
\]

\[
\text{Macro Pass Rate} =
\frac{\sum_{p \in P} \mathbf{1}_{\mathrm{passed}(C_p,p)}}
{|P|}.
\]

Intuitively, the micro score measures success \emph{constraint by constraint} across all plans, while the macro score is stricter, counting a plan only if it satisfies \emph{all} of its applicable constraints.

\paragraph{Final Pass Rate}
The proportion of plans that simultaneously (i) are delivered, (ii) pass all applicable commonsense constraints, and (iii) pass all applicable hard constraints.

%\clearpage

\section{Detailed Dataset Characteristics}
\label{app:dataset_table}
Table~\ref{tab:dataset_full} provides the comprehensive structural details of the TravelPlanner benchmark used in our evaluation, including trip lengths, difficulty distributions, and geographic coverage.

\begin{table}[H]
\centering
\caption{Comprehensive TravelPlanner Dataset Summary.}
\label{tab:dataset_full}
\begin{tabularx}{\linewidth}{l X}
\toprule
Total queries & 1,225 (train 45, val 180, test 1,000) \\
Trip lengths & 3, 5, 7 days \\
Difficulty & easy / medium / hard \\
Geographic spread & $\geq$192 origins, $\geq$127 destinations (test split) \\
Budgets (ours, for analysis) & \textless\$1K, \$1K--\$3K, \textgreater\$3K buckets \\
Environment & Closed-world sandbox with offline tools (no live APIs) \\
Licensing & CC BY 4.0 (dataset), MIT (code) \\
\bottomrule
\end{tabularx}
\end{table}

%\clearpage

\section{Model Hyperparameters}
\label{app:model_config}
To ensure reproducibility, the following parameters were used for all Large Language Model (LLM) calls within the framework:
\begin{itemize}
    \item \textbf{Temperature}: 0.7 for planning, 0.3 for evaluation.
    \item \textbf{Max Tokens}: 4,000 (planning), 1,000 (evaluation).
    \item \textbf{Top-p}: 0.9 (planning), 0.8 (evaluation).
    \item \textbf{Penalty Settings}: Frequency penalty: 0.1; Presence penalty: 0.1.
    \item \textbf{Refinement Limits}: Maximum of 5 iterations per plan with a 300-second global timeout.
\end{itemize}

%\clearpage

\section{Assessment Criteria}
\label{sec:assessment-criteria}
Each itinerary was evaluated using a {1-to-5 rating scale} based on how well it aligned with the {user profile} and {query}. The criteria for assigning ratings were:
\begin{enumerate}
    \item \textbf{Lowest Personalization}: The plan is {generic} or {completely misaligned} with the user’s profile and query. No effort is evident in catering to user preferences or constraints.
    \item \textbf{Poor Personalization}: Some elements loosely align with the user’s profile, but {most aspects are irrelevant} or mismatched. Minimal effort was made to personalize the itinerary.
    \item \textbf{Moderate Personalization}: The plan {partially} addresses user preferences and constraints. It includes both {relevant and irrelevant elements}, showing {some personalization but lacking depth}.
    \item \textbf{Good Personalization}: The plan {mostly aligns} with the user profile and query. It includes {tailored elements}, though {minor gaps or missed opportunities} exist.
    \item \textbf{Excellent Personalization}: The plan is {highly tailored} to the user's profile and query. There are {no significant misalignments}, and the plan demonstrates a {deep understanding of user preferences and constraints} with {highly relevant details}.
\end{enumerate}

%\clearpage

%\section{Detailed Ablation Study}
%\label{sec:ablation-full}

%o measure the contribution of individual components, we evaluated four variants of our framework under identical datasets and protocols. Table~\ref{tab:component_contrib_full} shows that removing either the Evaluator or Feedback Agent significantly degrades the \textit{Final Pass Rate}. The steepest decline occurs when both are removed (21\% vs. 52\% for the full system), underscoring the necessity of the modular generate-evaluate-refine loop.

%\begin{table}[ht]
%\centering
%\caption{Component Contribution Analysis Across All Metrics (Ablation).}
%\label{tab:component_contrib_full}
%\resizebox{\columnwidth}{!}{%
%\begin{tabular}{lccccc}
%\hline
%\textbf{Condition} & \textbf{Personalization} & \textbf{Final Pass} & \textbf{Hard Constraint} & \textbf{Commonsense} & \textbf{Delivery} \\
%& \textbf{Score (\%)} & \textbf{Rate (\%)} & \textbf{Pass Rate (\%)} & \textbf{Pass Rate (\%)} & \textbf{Rate (\%)} \\
%\hline
%\textbf{Full System} & \textbf{87} & \textbf{52} & \textbf{68} & \textbf{62} & \textbf{100} \\
%No Evaluator & 79 & 30 & 45 & 49 & 91 \\
%No Feedback Agent & 71 & 28 & 42 & 46 & 90 \\
%No Eval \& Feedback & 65 & 21 & 35 & 40 & 86 \\
%TravelPlanner Baseline & 56 & 10 & 28 & 15 & 89 \\
%\hline
%\end{tabular}%
%}
%\end{table}

%\clearpage

\section{Reproducibility Bundle}
\label{app:repro}

%\paragraph{Environment and Performance}
Experiments ran on Windows~10 (64-bit, build 10.0.26100) with Python 3.11 on x64 CPUs.
Planning proceeded for up to five iterations with a 300\,s timeout and default temperature 0.7 (seeds not fixed).
Measured end-to-end latency ranged from 30--300\,s depending on itinerary complexity, with typical sessions
involving 1--3 LLM calls per agent, 5--15 database queries, and 1.5k--7k tokens (0.5k--2k input; 1k--5k output),
at an estimated cost of USD~\$0.00015--\$0.0007 (GPT-4o).
While introducing communication/memory overhead compared to single-agent baselines,
the multi-agent design enabled modular specialization, parallelism, and feedback-driven refinement.

%\clearpage

\section{Extended Future Outlook}
\label{sec:future-outlook-full}

\subsection{Semantic Inter-Agent Protocols}
While current communication is limited to state passing, future iterations will introduce intent-aware messaging. A shared queryable memory will support knowledge transfer with provenance tracking, reducing redundancy in itinerary generation.

\subsection{Distributed Knowledge Sharing}
We plan to enable learning components that leverage distributed observations. Local heuristics will be aggregated into global summaries, using structured push-pull mechanisms with conflict resolution to improve personalization without repeated refinements.

\subsection{System Scalability}
To address potential bottlenecks in the centralized supervisor, we propose a distributed design. This involves session affinity enforced by load balancers and a Redis-backed stateless execution model, allowing the system to scale across containerized deployments.

\end{document}